\documentclass[conference]{IEEEtran}
\IEEEoverridecommandlockouts
\usepackage{cite}
\usepackage{amsmath,amssymb,amsfonts}
\usepackage{spconf}
\usepackage{epsfig}
\usepackage{caption}
\usepackage{subcaption}
\usepackage{algorithmic}
\usepackage{graphicx}
\usepackage{textcomp}
\usepackage{xcolor}
\usepackage[hidelinks]{hyperref}
\def\BibTeX{{\rm B\kern-.05em{\sc i\kern-.025em b}\kern-.08em
    T\kern-.1667em\lower.7ex\hbox{E}\kern-.125emX}}
\begin{document}

\title{An End-to-end food portion estimation framework \\ based on shape reconstruction from monocular image}

\name{Zeman Shao, Gautham Vinod, Jiangpeng He, Fengqing Zhu}
\address{School of Electrical and Computer Engineering, Purdue University, West Lafayette, Indiana, USA}

\maketitle

\begin{abstract}
Dietary assessment is a key contributor to monitoring health status. Existing self-report methods are tedious and time-consuming with substantial biases and errors. Image-based food portion estimation aims to estimate food energy values directly from food images, showing great potential for automated dietary assessment solutions. 
Existing image-based methods either use a single-view image or incorporate multi-view images and depth information to estimate the food energy, which either has limited performance or creates user burdens.
In this paper, we propose an end-to-end deep learning framework for food energy estimation from a monocular image through 3D shape reconstruction. We leverage a generative model to reconstruct the voxel representation of the food object from the input image to recover the missing 3D information.
Our method is evaluated on a publicly available food image dataset Nutrition5k, resulting a Mean Absolute Error (MAE) of 40.05 kCal and Mean Absolute Percentage Error (MAPE) of 11.47\% for food energy estimation.
Our method uses RGB image as the only input at the inference stage and achieves competitive results compared to the existing method requiring both RGB and depth information.
\end{abstract}


\section{Introduction}
\label{sec:intro}
Food is an essential component of everyday life and dietary assessment is a key contributor to monitoring human health status and preventing chronic diseases such as diabetes. However, it is difficult to assess accurate dietary intake due to the high complexity of the diet~\cite{Konif2021}. Conventional methods of self-report dietary assessment include food record, food frequency questionnaires, and 24HR recall~\cite{Thompson2017} that rely on trained nutrition analysts to estimate the dietary intake, which is either tedious and time-consuming or has substantial human biases and errors~\cite{Poslusna2009}. Our goal is to reduce the user burden through an automated approach to accurately assess dietary intake directly from the image of foods consumed at eating occasions.

Image-based dietary assessment address the challenges of automatic dietary intake monitoring from different aspects including food recognition, segmentation, and portion estimation~\cite{He_2021_ICCVW, mao2021_improved, Shao2021}. However, food portion estimation is a relatively under-explored area compared to other fields. Most existing methods focus on the use of single-view, multi-view, and depth information to determine the portion of the food from the image~\cite{gao2018food, fang_2016, jia2014accuracy}. Multi-view based methods require matching the correspondence between images using a 3D structure reconstruction process, which adds user burden and may not be practical in some applications. Depth-based methods require a depth sensor to capture the depth map which represents the pixel-wise distance from the object surface to the camera, but performance largely depends on the quality of the depth map. Most embedded consumer depth sensors do not provide sufficiently high resolution depth map at close range. Estimating food portion size from a single-view image requires the least amount of user input and device dependency, but is an ill-posed problem as the food energy is strongly correlated to the 3D structure of the food object while the 3D information is lost when the food objects are captured from 3D real-world space to 2D image space, which hinders further progress in this topic area. 

In this paper, we focus on the development of a novel end-to-end deep learning framework to estimate food portion size (in calories) from the monocular food image based on 3D voxel reconstruction. It consists of three modules:
\textbf{1) Shape reconstruction module} where we first adapt the conditional GAN architecture~\cite{cgan} to train a generative model to recover the missing 3D information from the input 2D food image by reconstructing the 3D shape of the food object.
\textbf{2) Food energy density regression module} where we use a deep regression model with a classification network as its backbone to estimate the food class probability distribution and the food energy density (in energy per volume) from the input image.
\textbf{3) food energy density adaption module} where we refine the estimated food volume with the adaptation of the class probability distribution and calculate the food energy by combing the estimated food energy density and the refined food volume.


We evaluate the proposed method on a large publicly available food image dataset, Nutrition5k dataset~\cite{thames2021nutrition5k}, which provides the groundtruth nutrition information for each image in the dataset.
Although our method only uses an RGB image as the input during the inference stage without any other additional information required, we achieve better results for food energy estimation compared to the prior art~\cite{thames2021nutrition5k}.  In Section~\ref{sec:ablation}, we further explore the backbone network and compare their experimental results. We observed that a better classification network can significantly improve food energy estimation performance, and we envision this would open doors to potential methods as improved backbone classification network will achieve better performance of food energy estimation in the future.

The main contribution of our paper can be summarized as follows:
\begin{itemize}
    \item We propose an end-to-end deep learning framework for food energy estimation from monocular images based on 3D shape reconstruction.
    \item The proposed method uses an RGB image as the only input at the inference stage and achieves competitive results compared to the method requiring both RGB and depth information.
    \item The proposed method with an improved backbone network achieves better performance, which would open doors to potentially more accurate food energy estimation.
\end{itemize}

\section{Related Work}
\label{sec:related}

\subsection{3D Shape Reconstruction}
Understanding the 3D structure of a scene from images is a fundamental problem in computer vision.
The spatial quantity of an object can be estimated given a full 3D reconstruction (for example, a dense point cloud) of the scene if the reference scale in the world coordinates is available. 
For 3D reconstruction, most existing works rely on multiple views of the scene using stereo vision techniques, or simultaneous localization and mapping (SLAM)~\cite{hartley2004, slam}.

3D scene reconstruction based on single-view images is an ill-posed problem since most 3D information has been lost during the projection from 3D world coordinates to the 2D pixel coordinates.
Existing work aims to reconstruct 3D structure of a general object from monocular images including learning to assign geometric labels to superpixels~\cite{hoiem05}, and depth estimations based on Convolutional Neural Networks (CNNs)~\cite{casser2019unsupervised}.
We propose to learn 3D structure of foods in an image represented by voxel of objects in 3D space using the deep learning-based method.

\subsection{Image-based Food Portion Estimation}
Image-based food energy estimation aims to develop efficient processes that can automate the assessment of food energy from the eating occasion image directly and to alleviate some of the burden placed on dietitians and participants.
Deep learning techniques that utilize Convolutional Neural Networks (CNN) have achieved impressive results on many computer vision problems that aim to understand the visual world and objects within and they have also been used in food energy estimation.
Two methods that incorporate volume scalar or depth are proposed to estimate the food nutrient value from a single image~\cite{thames2021nutrition5k}.
The volume scalar-based method requires a volume scalar inferred from the captured depth map for direct regression estimation. The depth-based method augments the input image with depth as a 4th channel and uses it as input to the regression network.
However, both methods require additional information as input to estimate the food energy, which adds to the user burden in real-world applications. 
Previously, a generative network trained on the cGAN has been developed to generate the food energy distribution map which models the characteristics of energy distribution in a food image~\cite{fang2018}. 
The estimated energy distribution map is used as an intermediate result to a regression network to further estimate the food energy~\cite{fang2019end}, augmenting with the features extracted from the RGB image shows an improvement of the food energy estimation~\cite{Shao2021}.
A method that incorporates depth information for joint learning is proposed in~\cite{vinod2022} and is evaluated on a subset of Nutrition5k dataset~\cite{thames2021nutrition5k} which is a large public food image dataset with  depth map and groundtruth nutrition information.

\begin{figure*}[t]

    \centering
    \includegraphics[width=\textwidth]{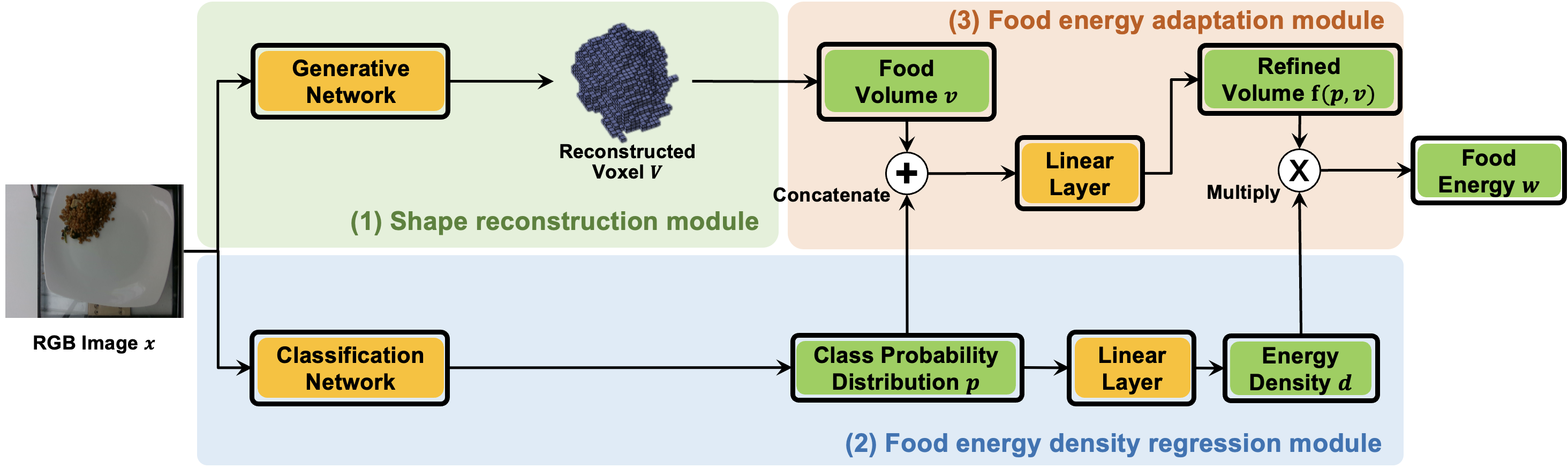}
    \caption{
    \textbf{Overview of proposed end-to-end framework to estimate food energy}. 
    The shape reconstruction module reconstructs the food object voxel by a generative model. 
    The Food energy density regression module estimates the food energy density (in energy per volume) from the input food image.
    The food energy density adaption module determines the food energy based on the estimated food energy density and food object voxel.
    }
    \label{fig:diagram}
\end{figure*}

\section{Proposed Method}
\label{sec:method}
Food energy is a single numerical value that is directly related to both the type of food object and its volume referred to as the spatial quantity in 3D space. However, most 3D information is lost when the food object is captured by cameras from the 3D world space to the 2D image space, making it difficult to obtain the corresponding volume size. To address this challenge, we propose an end-to-end deep learning framework to estimate food energy from the monocular food image through 3D shape reconstruction.

The overview of our proposed framework is shown in Figure~\ref{fig:diagram}, which consists of the following three modules:
\begin{enumerate}
    \item \textbf{Module 1: 3D Shape reconstruction module} aims to recover the 3D shape of the food object based on the input 2D monocular image by a generative model.
    \item \textbf{Module 2: Food energy density regression module} estimates food class probability distribution and the energy density (in the unit of energy per volume) from the food image.
    \item \textbf{Module 3: Food energy adaptation module} is used to refine the estimated food volume with the adaptation of the class probability distribution and the refined food volume is then multiplied by the estimated energy density to calculate the food energy.
\end{enumerate}

\subsection{3D Shape Reconstruction Module}
\label{sec:method-gan}
The objective of this module is to obtain the lost 3D information by reconstructing the 3D shape given the 2D image of food objects by applying Conditional GAN (cGAN)~\cite{cgan} based on an input image $x$. Specifically, we leverage pix2pix framework~\cite{pix2pix} with a 3D U-net~\cite{knyaz2018zgan} as the generative model to compute the 3D shape of the food object represented by the voxel $V$, which is in the form of 3D unit grids with pre-defined volume to represent a 3D model. The 3D U-net has the same encoder network and skip connections as the U-net, but the 2D deconvolution layers are replaced by the 3D deconvolution layer in order to decode the correlation between neighboring voxels along the Z-axis~\cite{knyaz2018zgan}. Thus, the slice of the reconstructed voxel $V$ along the Z-axis is supposed to be aligned with the contour of the input image. 

To train the cGAN, the voxel representation of the food object in the corresponding image is required. 
The raw depth map $D$ represents the pixel-wise distance from the object surface to the camera, and it can be used to generate the voxel representation $V$ of the object in the image.
However, the raw depth map inevitably contains missing pixels if the surface is shiny, transparent, too close, or too far, and it also includes the background region with no food objects.
Therefore, We need to inpaint the missing parts of the raw depth map and remove the non-food region before we use it to generate the food object voxel.
Following the depth post-processing described in~\cite{vinod2022}, we performed the depth map dilation for the depth completion and used the food segmentation mask estimated by Google’s Seefood
mobile food segmenter~\footnotetext[1]{The code of Google's Seefood mobile food segmentation model is available at \url{https://tfhub.dev/google/seefood/segmenter/mobile\_food\_segmenter\_V1/1}} to eliminate the non-food background region, and obtained the normalized depth map $\Bar{D}$. 

\begin{figure*}[t]
     \centering
     \begin{subfigure}[b]{0.26\textwidth}
         \centering
         \includegraphics[width=\textwidth]{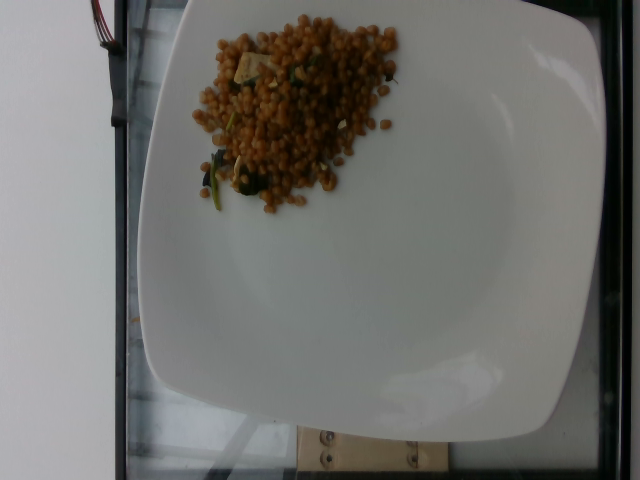}
         \caption{Dish image $x$}
          \label{fig:rgb}
     \end{subfigure}
          \hfill
     \begin{subfigure}[b]{0.26\textwidth}
         \centering
         \includegraphics[width=\textwidth]{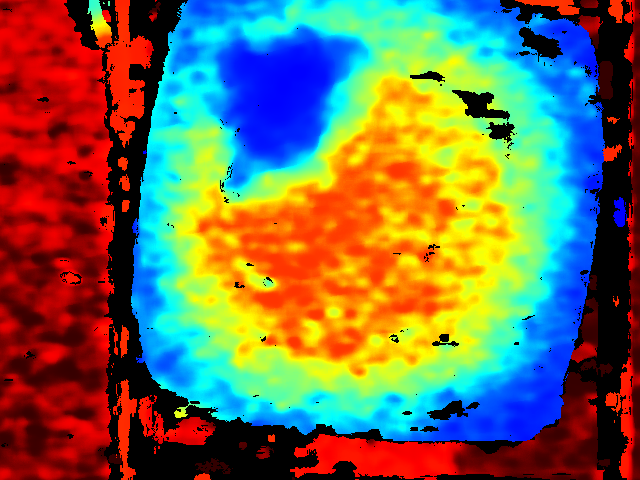}
         \caption{Raw depth map $D$}
          \label{fig:raw-depth}
     \end{subfigure}
     \hfill
     \begin{subfigure}[b]{0.26\textwidth}
         \centering
         \includegraphics[width=\textwidth]{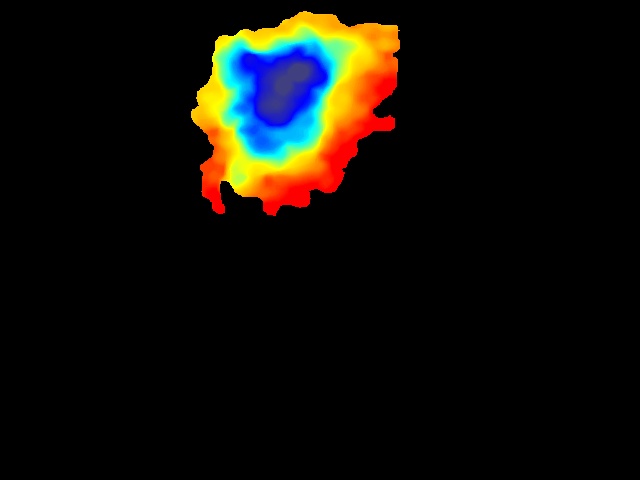}
         \caption{Post-processed depth map $\Bar{D}$}
          \label{fig:post-depth}
     \end{subfigure}
     \hfill
     \begin{subfigure}[b]{0.20\textwidth}
         \centering
         \includegraphics[width=\textwidth]{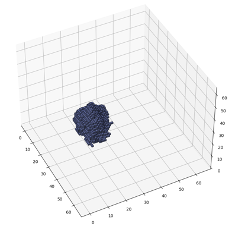}
         \caption{Reconstructed voxel $V$}
          \label{fig:vox}
     \end{subfigure}
        \caption{An example dish image from Nutrition5K dataset. (a) RGB image, (b) corresponding raw depth map (c) corresponding depth map after post-processing (d) reconstructed voxel representation of food object. The associated groundtruth food energy is 92.41 kCal. (Depth maps shown above are colorized for visualization)}
        \label{fig:ex}
\end{figure*}

The post-processed depth map $\Bar{D}$ is converted into a voxel representation where we categorize each voxel grid as free or occupied one, depending on whether it is in front of or behind the visible surface from the depth map. It can be expressed as:
\begin{equation}
     V(x, y, z) = 
     \begin{cases} 
      1 & z\geq \Bar{D}(x, y) \\
      0 & z < \Bar{D}(x, y)
   \end{cases}
     \label{equ:vox}
\end{equation}
Following Equation~\ref{equ:vox}, we generate the food object voxel based on the post-processed depth map $\Bar{D}$. The generated voxels are used as the reference to train the cGAN in the shape reconstruction module as illustrated in Section\ref{sec:method-gan}.
An example of the original RGB image (Figure~\ref{fig:rgb}), the depth map of the food object (Figure~\ref{fig:post-depth}) and its associated voxel of the food object (Figure~\ref{fig:vox}) is shown in Figure~\ref{fig:ex}, the depth map is colorized for visualization purpose only. 

Therefore, the cGAN in this module learns the mapping from a random noise vector $z$ to a target voxel $V$ conditioned on the observed image $x$. 

\subsection{Food Energy Density Module}
\label{sec:method-density}
The food energy density $d$ (in the unit of energy per volume) is the property determined by the food types, thus estimating the food energy density from the image can be regarded as an image classification task. As a deep neural network has shown impressive results on the food image classification~\cite{He_2021_ICCVW, mao2020visual}, it is sufficient for learning the food energy density value directly from the input images. Therefore, in this food energy density module, we apply an off-the-shelf classification network as the backbone to perform food classification and to estimate food energy density based on predicted food types. Specifically, we attached additional linear layers to the original classification network to output a single value as the predicted food energy density. In addition, we explored popular deep learning models as the backbone network for food classification including VGG network~\cite{vgg}, ResNet~\cite{resnet}, DenseNet~\cite{densenet} and Swin Transformer~\cite{swin}. Due to the few available training data, we leverage the pre-trained weights on ImageNet~\cite{imagenet} for the classification network.
We random initialize the attached linear layers and fine-tune the weights of the classification network for the food density estimation task based on the building blocks of complex features originally learned from ImageNet~\cite{imagenet}. With the food energy density module, we can estimate the food energy density $d$ in the unit of energy per volume from one monocular RGB image.

\subsection{Food Energy Adaption Module}
\label{sec:method-energy}
Once we obtained the voxel representation $V$ of the food object from the input RGB image using the generative network trained by the cGAN as illustrated in Section~\ref{sec:method-gan}, we can then determine the volume $v$ for the food object in the image as 
\begin{equation}
    v = \sum_{x,y,z}~V(x,y,z)    
\end{equation}
However, the above-obtained food volume depends solely on the reconstructed voxel where potential error from the estimated voxel representation of the food object could propagate into the subsequent food energy estimation.
Therefore, we propose to refine the food volume obtained from the estimated voxel in module 1 with the adaptation of the features from the image classification network in module 2.
The output of the probability distribution vector from the image classification network represents the probability of all estimated food classes. It is used as an additional feature to refine the estimated food volume.
We denote the probability distribution vector from the classification network as $p$, and the estimated food volume as $v$.
We first concatenate the $p$ and $v$ and then apply a linear layer to obtain the refined volume.
The estimated food energy $w$ can be calculated by multiplying the refined volume with the estimated energy density from module 2, which is expressed as
\begin{equation}
    w = d \times f(p, v)
\end{equation}
where $f(\cdot)$ is the linear function to obtain the refined volume, and $d$ is the food energy density (in the unit of energy per volume) estimated in module 2.

\section{Experimental Results}
\label{sec:results}
In this section, we evaluate the proposed method by comparing it with existing work and also conduct an ablation study to show the effectiveness of each proposed module introduced in Section~\ref{sec:method}. 

\textbf{Dataset:} we use the publicly available Nutrition5k dataset~\cite{thames2021nutrition5k}, which contains short videos and images of around 5,006 unique dishes constructed from more than 250 different ingredients. Each dish has a breakdown of ingredient names, and associated macronutrient information. We use the 3,259 top-view food dish image from the Nutrition5k dataset since only the top-view images have the associated depth map. We used the post-processed depth map to generate the food object voxel which is used as the refernece to train the cGan in the shape reconstruction module as illustrated in Section\ref{sec:method-gan}. 
An example of the top view food dish image (Figure~\ref{fig:rgb}), its associated raw depth map (Figure~\ref{fig:raw-depth}), the depth map after post-processing (Figure~\ref{fig:post-depth}) and its associated voxel of the food object (Figure~\ref{fig:vox}) are shown in Figure~\ref{fig:ex}, the depth map is colorized for visualization purpose only.
We adopt the same data split used in~\cite{thames2021nutrition5k} to evaluate our method for fair comparison where a total of 2,753 images were used to train the framework and 506 images for testing. 

\textbf{Evaluation metrics:} We used two common regression metrics, mean absolute error (MAE) and mean absolute percentage error (MAPE), defined as
\begin{equation}
\text{MAE} = \frac{1}{N}\sum^{N}_{i=1} |\tilde{w}_i - \bar{w}_i|
\end{equation}
\begin{equation}
\text{MAPE} = \frac{100\%}{N}\sum^{N}_{i=1} \frac{|\tilde{w}_i - \bar{w}_i|}{\bar{w}_i}
\end{equation}
where $\tilde{w}_i$ is the estimated food energy of the $i$-th image, $\bar{w}_i$ is the groundtruth food energy of $i$-th image and $N$ is the number of testing images. The MAE represents the mean absolute difference between the actual and the predicted value, while the MAPE reflects the mean absolute percentage difference between the actual and the predicted value.
The MAE and MAPE reported below are calculated based on the models after 50 training epochs. 

\subsection{Comparison to Existing Methods}
We compared our method to two other methods introduced in~\cite{thames2021nutrition5k}. In addition to the RGB input, the volume scalar method requires a volume scalar inferred from the captured depth map as an additional input for regression module during both training and inference stage. The depth as 4th channel method augments the input image with depth as a 4th channel, and is used as input to the regression network for both training and inference stage. On the contrary, our method only uses the depth map to generate the reference food object voxel for training the cGAN at the training stage. 
Once the generative model is trained, the RGB image is the only input required at the inference stage. Table~\ref{tab:comprasion} shows the results for these three methods, where our proposed method outperforms both baseline methods without requiring depth information during inference. Note that, the metric, MAE over Mean (MAEoM), is the mean absolute error as the percentage of the mean groundtruth energy which is introduced in~\cite{thames2021nutrition5k}.


\begin{table}[t]
\fontsize{9}{11}\selectfont
\begin{center}
\caption{Experimental results for food energy estimation on nutrition5k dataset compared to \cite{thames2021nutrition5k}}
\label{tab:comprasion}
\begin{tabular}{*5c}
\hline
{} & Inference Input  & MAE (kCal)   & MAEoM (\%) \\
\hline\hline
\textbf{Ours}      &   \textbf{RGB}    &  \textbf{40.05} & \textbf{15.8}\\
\hline\hline
~\cite{thames2021nutrition5k}   & RGB + Volume Scaler          & 41.30  & 16.5\\
~\cite{thames2021nutrition5k}   & RGB + Depth  & 47.6  & 18.8\\
\hline
\end{tabular}
\end{center}
\end{table}

\subsection{Ablation Analysis}
\label{sec:ablation}
We conducted an ablative analysis to compare different backbone classification network in the food energy density regression module. We also investigated the influence of different modules in the proposed framework. 

\textbf{Backbone network:}
To compare different backbone classification networks in the food energy density estimation module, we explored the usage of VGG-16, ResNet-50, DenseNet-201, and Swin Transformer-base~\cite{vgg, resnet,densenet, swin} in the food energy density estimation module.
We summarize the results in Table~\ref{tab:backbone}.
The two CNN-based networks with deeper architecture, DenseNet and ResNet, achieved a MAE of 49.92 kCal and 46.23 kCal and a MAPE of 26.3\% and 23.3\%, respectively, showing better performance than VGG where the MAE and MAPE are 63.69 and 33.0\%, respectively.
Next, using the Swin Transformer as the backbone network achieved even better results both in MAE and MAPE, 40.05 kCal and 22.0\%, respectively and as expected.
We observe that a better network model for image classification can improve the food energy estimation accuracy in our proposed framework.
\begin{table}[t]
\fontsize{9}{11}\selectfont
\begin{center}
\caption{Experimental results for food energy estimation using different classification backbone networks}
\label{tab:backbone}
\begin{tabular}{*5c}
\hline
Backbone Network in Module 2  & MAE (kCal)   & MAPE (\%) \\
\hline
VGG                                 & 63.69  & 33.0\\
DenseNet                            & 49.92  & 26.3\\ 
ResNet                              & 46.23  & 23.3\\
\textbf{Swin Transformer}           & \textbf{40.05} & \textbf{22.0}\\                                            
\hline
\end{tabular}
\end{center}
\end{table}

\textbf{Modules in the framework:}
In this experiment, we fixed the Swin Transformer as the backbone classification and explored the influence of different modules in the proposed framework. Results are summarized in Table~\ref{tab:ablation}. We use module 2 of our method as the baseline method where the classification network with one additional linear layer is used to estimate the food energy directly.
We achieved an MAE of 51.26 kCal and MAPE of 26.98\% using the baseline. If we directly use the volume estimated from the reconstructed voxel (module 1) without any refinement from module 3, there is a performance degradation in both MAE and MAPE, 76.74 kCal and 28.44\%, respectively. 
The reconstructed voxel inevitably has false detection, such as miss-detected voxel grids. Thus, these errors can propagate into the food volume and energy calculation.

\begin{table}[t]
\fontsize{9}{11}\selectfont
\begin{center}
\caption{Experimental results for food energy estimation on nutrition5k}
\label{tab:ablation}
\begin{tabular}{*5c}
\hline
Method  & MAE (kCal)   & MAPE (\%) \\
\hline\hline
Module 2                                 & 51.26  & 26.98\\
Module 1 and 2                           & 76.74  & 28.44\\
\textbf{Module 1-3 (Proposed method)}    & \textbf{40.05} & \textbf{22.0}\\ 
\hline
\end{tabular}
\end{center}
\end{table}

\section{Conclusion}
\label{sec:conclusion}
In this paper, we proposed a deep learning framework to estimate food energy from monocular images based on 3D shape reconstruction. 
Our method required RGB image as the only input, and achieved promising performance of food energy estimation on the top-view food image from the Nutrition5k dataset.
The method with an improved backbone classification network achieved a better performance of food energy estimation, which shows great potential for advancing the field of image-based dietary assessment.
Our future work will focus on combining physical reference and contextual dietary information
to further improve the accuracy of the end-to-end food portion estimation.
\bibliographystyle{IEEEbib}
\bibliography{icme2023template}

\end{document}